\pdfoutput=1
\documentclass[conference]{IEEEtran}
\IEEEoverridecommandlockouts
\usepackage{cite}
\usepackage{amsmath,amssymb,amsfonts}
\usepackage{algorithmic}
\usepackage{graphicx,subcaption}
\usepackage{textcomp}
\usepackage{xcolor}
\newcommand{\concat}{\ensuremath{+\!\!\!\!+\,}}
\def\BibTeX{{\rm B\kern-.05em{\sc i\kern-.025em b}\kern-.08em
T\kern-.1667em\lower.7ex\hbox{E}\kern-.125emX}}
\begin{document}

\title{Learning Wear Patterns on Footwear Outsoles Using Convolutional Neural Networks}

\author{
\IEEEauthorblockN{
Xavier Francis\IEEEauthorrefmark{1}, Hamid Sharifzadeh\IEEEauthorrefmark{1}, Angus Newton\IEEEauthorrefmark{2}, Nilufar Baghaei\IEEEauthorrefmark{3}, and Soheil Varastehpour\IEEEauthorrefmark{1}\\[8pt]}

\IEEEauthorblockA{
  \IEEEauthorrefmark{1}School of Computing\\
  Unitec Institute of Technology\\
  Auckland, New Zealand\\
  \{xfrancis, hsharifzadeh, spour\}@unitec.ac.nz\\[8pt]}

\IEEEauthorblockA{
  \IEEEauthorrefmark{2}Physical Evidence Team\\
  Institute of Environmental Science and Research (ESR)\\
  Auckland, New Zealand\\
  angus.newton@esr.cri.nz\\[8pt]}

\IEEEauthorblockA{
  \IEEEauthorrefmark{3}Department of Information Technology\\
  Otago Polytechnic Auckland Campus (OPAIC)\\
  Auckland, New Zealand\\
  nilufar.baghaei@op.ac.nz}}

\maketitle

\begin{abstract}
Footwear outsoles acquire characteristics unique to the individual wearing them over time. Forensic scientists largely rely on their skills and knowledge---gained through years of experience---to analyse such characteristics on a shoeprint. In this work, we present a deep learning model that, for the first time, can predict the wear pattern on a unique dataset of shoeprints that captures the life and wear of a pair of shoes. We also present an additional architecture able to reconstruct the outsole back to its original state on a given week, and provide empirical evaluations of the performance of both models.
\end{abstract}

\begin{IEEEkeywords}
Shoeprints, Deep Learning, CNN, Predictive Modeling, Wear-and-Tear
\end{IEEEkeywords}

\section{Introduction}
Among the many forms of physical evidence found at crime scenes, shoeprints are one of the most frequently seen, with a high degree of evidential value attributed to them. Marks and prints formed by the footwear worn by the criminal(s) are frequently found at scenes-of-crime and their study was being recorded as early as 1786 \cite{LIUKKONEN199699}. This is in part due to the ability of a shoeprint to uniquely identify an individual, by evaluation of the combination of tears, nicks, cuts, scratches and other abrasions that form on the outsole as a function of wear. This `wear pattern' is influenced by biomechanics such as the weight and gait of the wearer, enviromental stressors, and additional factors like the material of construction. Bodziak defined wear as ``the erosion of the outsole due to abrasive forces that occur between the outsole and the ground'' \cite{bodziak1999footwear}. By considering the wear pattern, in addition to the pattern of the outsole introduced in the manufacturing process, one is able to ascertain if the shoe of a suspect formed the print found at the crime scene. 

Inspite of their uniquely identifiable nature and their frequency of appearance at scenes-of-crime, shoeprints are not often used as evidence in a court of law. This is in part due to the variation in quality of scene-of-crime impressions, which are often incomplete or degraded. Another challenge is the large search space of potential outsoles; arising from the number of outsoles being designed and manufactured. 

Consider the scenario where a substantial period of time elapses between the perpetration of a crime and the identification of suspect(s). In such situations, it falls on the forensic scientist to evaluate the outsole and determine if it matches the scene print while accounting for the formation of additional wear features. This task involves the careful analysis of the outsole and requires intimate knowledge of the breadth of factors and variables that influence wear patterns.

The forensic examiner's interpretation of the shoeprint and its admissability as evidence is built through their years of experience in studying shoeprints and the individualising characteristics that contribute to the wear pattern. Such knowledge is notoriously hard to quantify and explain. Deep learning models have made large strides in developing representations of domains like these. 

\begin{figure*}[ht]
  \centering
  \begin{minipage}[t]{\textwidth}
  \begin{subfigure}[t]{0.3\textwidth}
      \includegraphics[width=\textwidth]{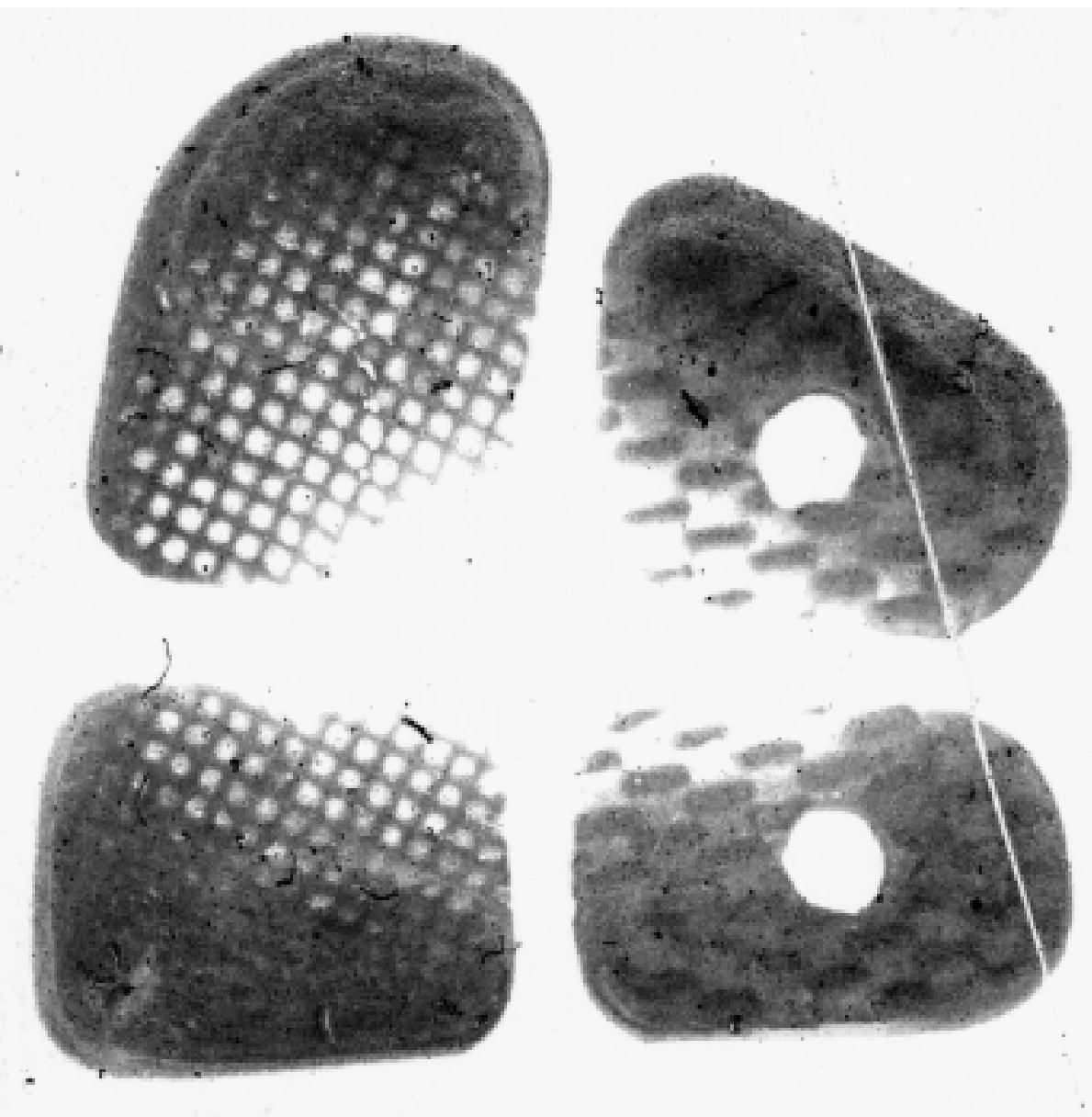}
      \caption{Cropped region of heel on week 4. Noise is evident.}
      \label{fig:04L-crop}
  \end{subfigure}
  \hfill
  \begin{subfigure}[t]{0.3\textwidth}
    \includegraphics[width=\textwidth]{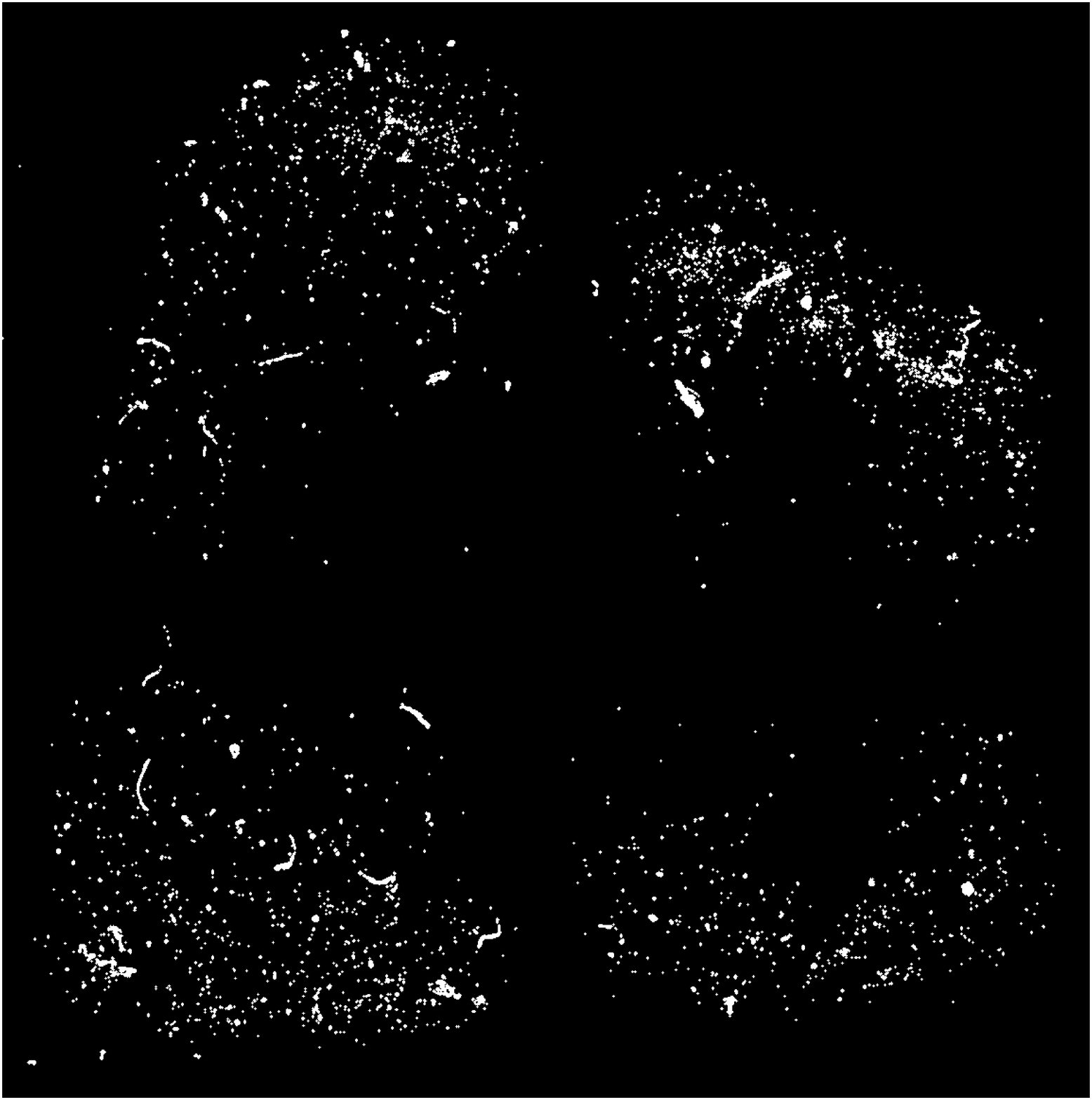}
    \caption{Noise map of image \ref{fig:04L-crop} obtained via thresholding.}
    \label{fig:04L-nm-dilated}
  \end{subfigure}
  \hfill
  \begin{subfigure}[t]{0.3\textwidth}
      \includegraphics[width=\textwidth]{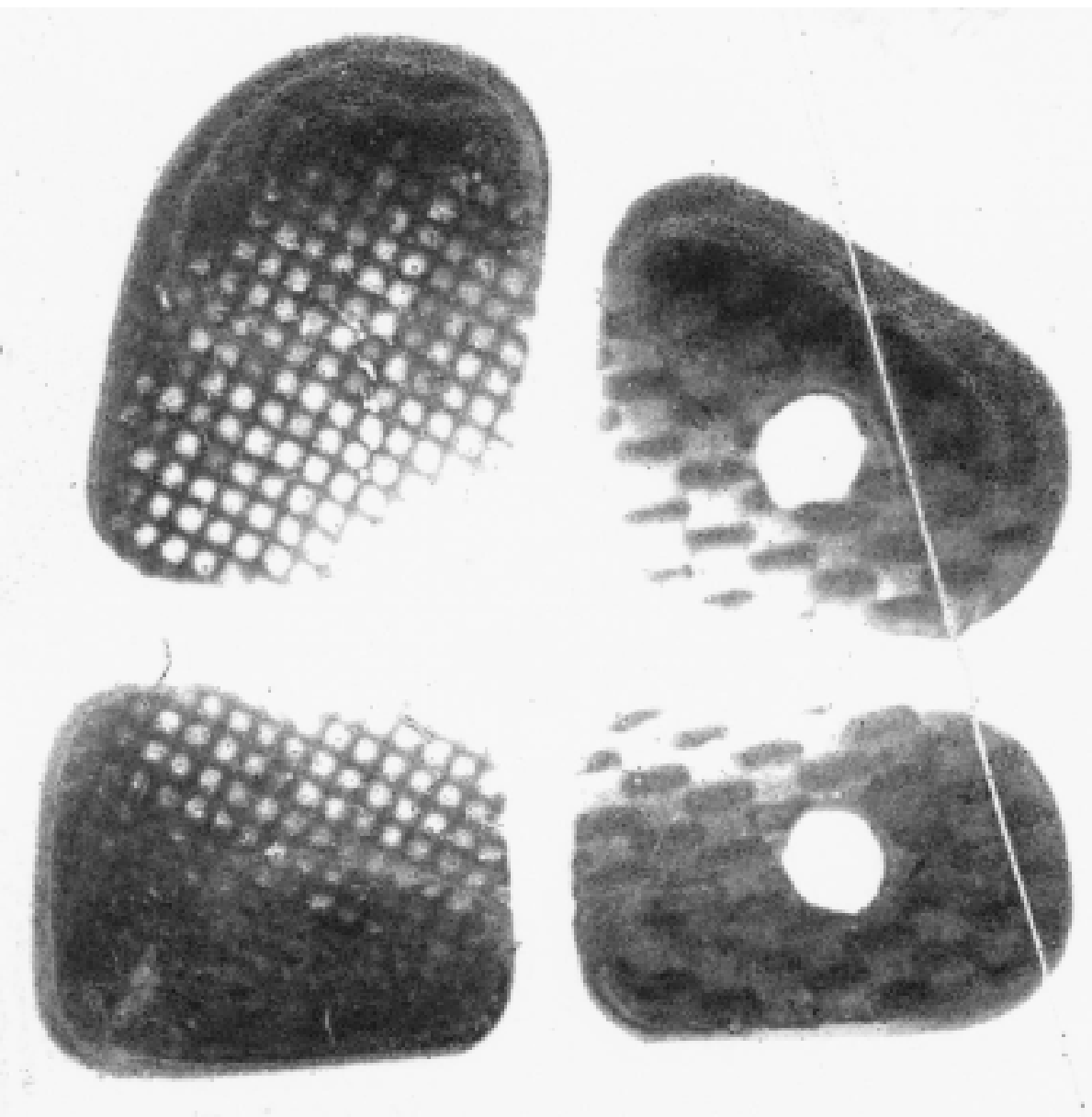}
      \caption{Fully filtered image, showing mitigated noise.}
      \label{fig:04L-filtered}
  \end{subfigure}
  \caption{Intermediate stages of the denoising methodology developed for our dataset.}\label{fig:denoise}
\end{minipage}
\end{figure*}

In this work, we adapt a convolutional neural network (CNN) architecture for the task of pixel-wise prediction of shoeprint wear. Our core contributions are as follows---(i) we describe a methodology that utilises a CNN to predict outsole wear formation on a unique dataset of shoeprints, and (ii) an alternate architecture that is able to reconstruct the outsole back to its original state on a given week within a timeframe of one year. 

In the following sections we first survey the related literature in the domains of forensics and shoeprints, and deep learning; followed by a description of our novel dataset. We proceed to detail our methodology, analyse the results of our experiments, and finally we conclude the paper. 

\section{Related Work}
\subsection{Shoeprint Classification}
The responsibilities of the forensic footwear examiner are: (i) to identify the make and model of a given shoeprint, by comparing it against a large set of known prints and (ii) to consider the individualising characteristics of the print to assign the print to an owner. The first task is largely objective in nature; by comparing the scene-of-crime print against a database of reference shoeprints, one is able to find a match and retrieve the relevant metadata. Numerous computational methods have been developed over the years to assist the examiner in this task. 

Automated approaches to shoeprint retrieval and classification have seen a multitude of approaches --- Fourier features \cite{GERADTS199621}, fractals \cite{791138}, power spectral density \cite{1388261}, Hu moment invariants \cite{ALGARNI200810}, Harris points and SIFT descriptors \cite{5319317}, Mahalanobis distance as feature descriptors \cite{DardiFootwear2009}, wavelets as an edge detector and neural networks for recognition \cite{wang2009research}, and transforms like Radon and Gabor \cite{PATIL20091308} \cite{5304124}.

In 2017, Richetelli et al. \cite{RICHETELLI2017102} postulated that the recent advances in deep learning could carry over to field of footwear classification. Kong et al. \cite{kongcross} and Zhang et al. \cite{10.1007/978-3-319-69923-3_56} were some of the first to apply CNNs to this task. However, they have not considered wear patterns and our work can be seen as a new contribution to the literature on deep learning applications in forensic science.

\subsection{Shoeprint Wear}
While the above research considers the challenge of using computational methods to aid in the task of shoeprint identification, our focus is on using computational methods to model shoeprint wear; specifically, we consider how outsole features change over time. Research in this domain is sparse, with a few considering wear formation manually \cite{wyatt2005aging} \cite{adair2007mount}, and fewer using pattern recognition techniques \cite{JFO:JFO1209} \cite{SHEETS201384}. All of the above mentioned studies vary in scale, time, and ambition. Understandably, controlling the variables that influence outsole wear is in itself a challenge.

\subsection{Image-to-Image Regression}
Given our dataset of 52 shoeprints, described in \ref{sec:dataset}, we wish to learn a model of the wear pattern captured within. Once trained, this model should be capable of extrapolating the wear pattern on seeing a new shoeprint. Fundamentally, we approach this as an image-to-image regression task. The literature contains many successful applications of deep learning to these types of dense prediction tasks; such as image in-painting \cite{pathakCVPR16context} \cite{yu2018generative}, super-resolution \cite{DongImage2014}, denoising \cite{Zhang2017DNCNN}, and image recovery from compressed representations \cite{dong2015compression}. Deep neural networks (DNNs) and their convolutional variants have established state-of-the-art performance over nearly all facets of computer vision tasks. One of the primary advantages of using DNNs is their ability to learn end-to-end mappings without the use of image priors, or the explicit engineering of features.

Our dataset shows the life and wear of a pair of shoes through impressions captured at evenly spaced intervals of time. To the best of our knowledge, this is the first time such a dataset has been used in the literature of deep learning. A closely related problem is video frame generation/prediction \cite{mathieu2015deep} that involves operations on inputs in the spatial domain, while simultaneously capturing correlations in the temporal domain. Notably, in video frame prediction, one has access to an extensive amount of data by using each frame in the video sequence as a datapoint. Finn et al. \cite{Finn2016Unsupervised} use a combination of convolutional and LSTM layers to model pixel motion and optical flow. They introduce a dataset with 1.5 million video frames and a model that predicts video sequences up to 1 second in the future. Our dataset, described in the next section, is significantly smaller in size.

\begin{figure*}[ht]
  \centering
  \includegraphics[width=\textwidth,keepaspectratio]{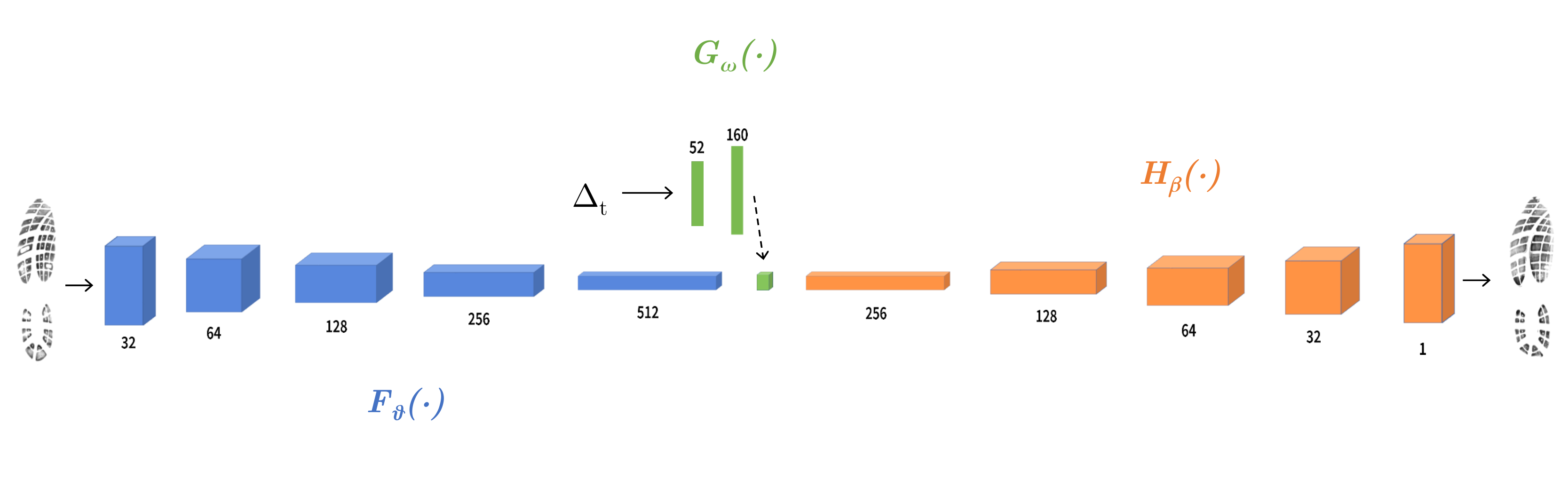}
  \caption{A visualisation of the architecture of our CNN.}\label{fig:arch}
\end{figure*}

\section{Dataset}
\label{sec:dataset}
For the collection of our dataset, we limit the influence of environmental variables and consider the formation of wear characteristics on a single pair of shoes, worn daily by one individual, over the course of one year. A pair of \textit{Asics}-brand men's sneakers were purchased and worn by the forensic specialist everyday for a period of 52 weeks in an urban environment. Impressions of both outsoles were then captured every fortnight using \textit{BVDA} gels. These impressions were scanned into high-resolution digital negative \textit{TIFF} files, yielding a total of 52 files, including week 0 (unfortunately, impressions were not captured on week 6). Each file is a 256-level grayscale image.

The outsole itself consists of approximately 63 `block features' of varying size and shape. Some of these block features contain other features within --- such as the \textit{Asics} brand logo, which is is not visible in the impression from week 0, but reveals itself over time; a text pattern that says \textit{`GEL'}; and circular `holes' present in many of the block features. These are all patterns imprinted by the manufacturer and their appearance and disappearance can be observed as they wear. 

During the course of recording these impressions, many forms of unwanted features were captured in addition to the shoeprint. These included air bubbles, fingerprints, dust, debris overlapping the shoeprint, ghosting of the impression, and areas of missing detail. Of these, the most egregious was determined to be the debris which appeared to consist of fibres and other objects that were transferred from the outsole and onto the gel in the process of imprinting. The debris was particularly problematic due to it obscuring regions of interest in the shoeprint. We refer to these unwanted features that obscured the object of interest as `noise'. 

To address this, we developed a denoising method capable of maintaining the high and low-level wear patterns, while simultaneously mitigating the noise present in the dataset. This methodology operates by using a local adaptive thresholding to obtain a binary mask, ROI filtering and morphological operations to process the mask, creating a noise map which allows for processing each block feature of the outsole independently, and using an averaging filter to mitigate the noise. Figure \ref{fig:denoise} highlights a few stages of this denoising methodology. The shoeprint images were cropped, denoised, and registered in the data preparation stage. 

Next, we articulate the CNN architectures developed to model the wear pattern on this dataset.

\begin{figure*}[ht]
  \centering
  \begin{minipage}[t]{\textwidth}
  \begin{subfigure}[t]{0.23\textwidth}
      \includegraphics[width=\textwidth]{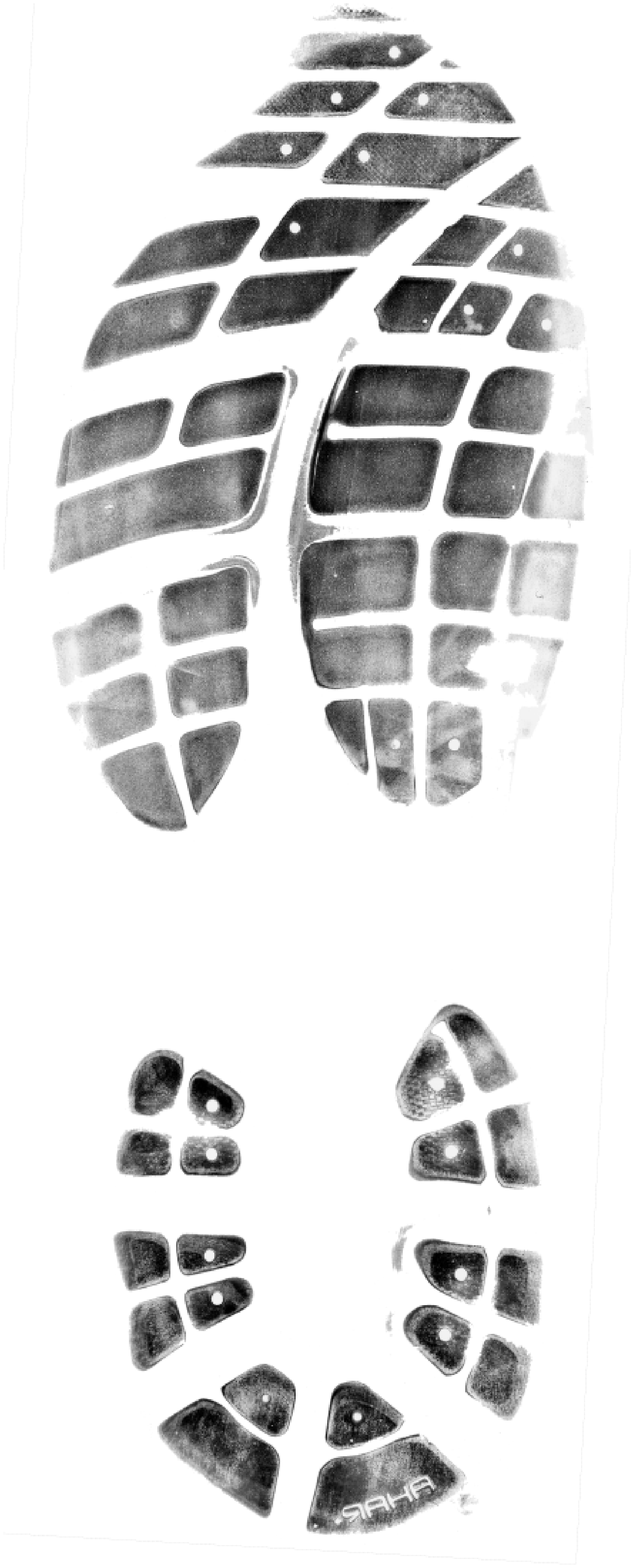}
      \caption{Input image of left outsole on week 48.}
      \label{fig:48L-input}
  \end{subfigure}
  \hfill
  \begin{subfigure}[t]{0.23\textwidth}
      \includegraphics[width=\textwidth]{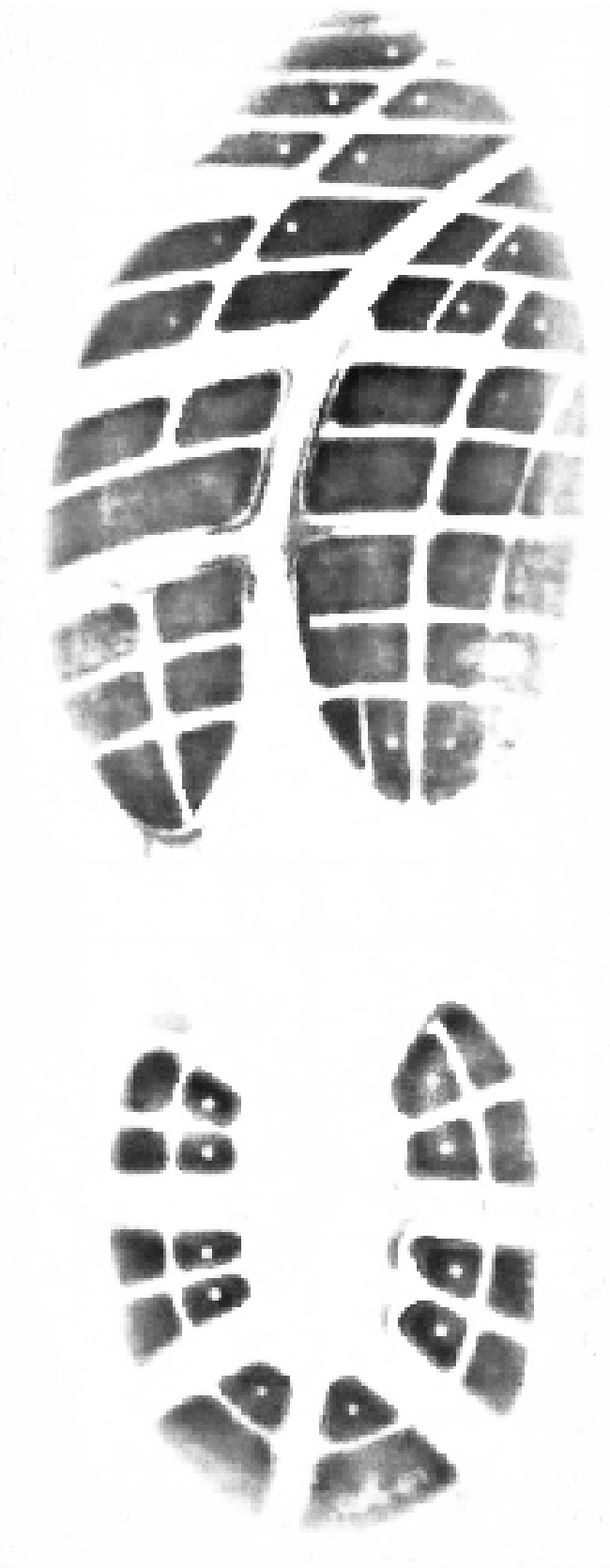}
      \caption{$\Delta t = 8$.}
      \label{fig:8-pred}
  \end{subfigure}
  \hfill
  \begin{subfigure}[t]{0.23\textwidth}
    \includegraphics[width=\textwidth]{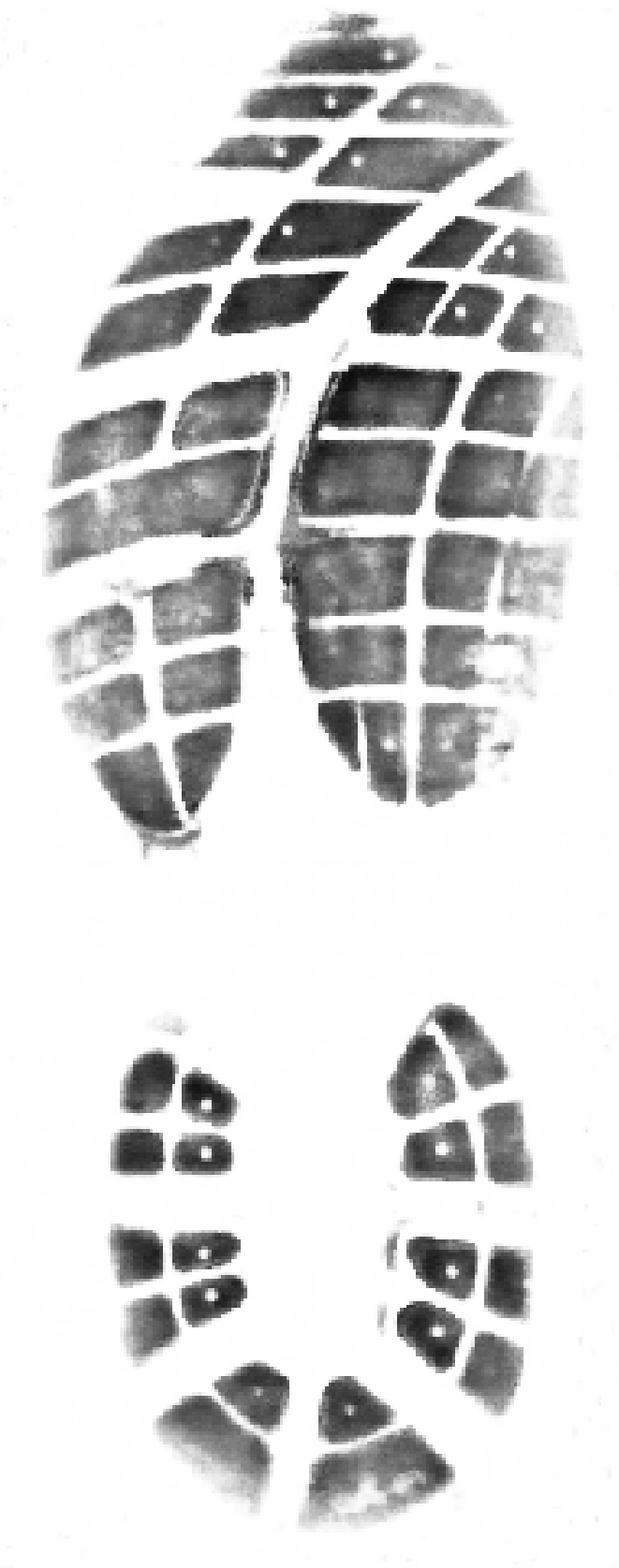}
    \caption{$\Delta t = 24$.}
    \label{fig:24-pred}
  \end{subfigure}
  \hfill
  \begin{subfigure}[t]{0.23\textwidth}
      \includegraphics[width=\textwidth]{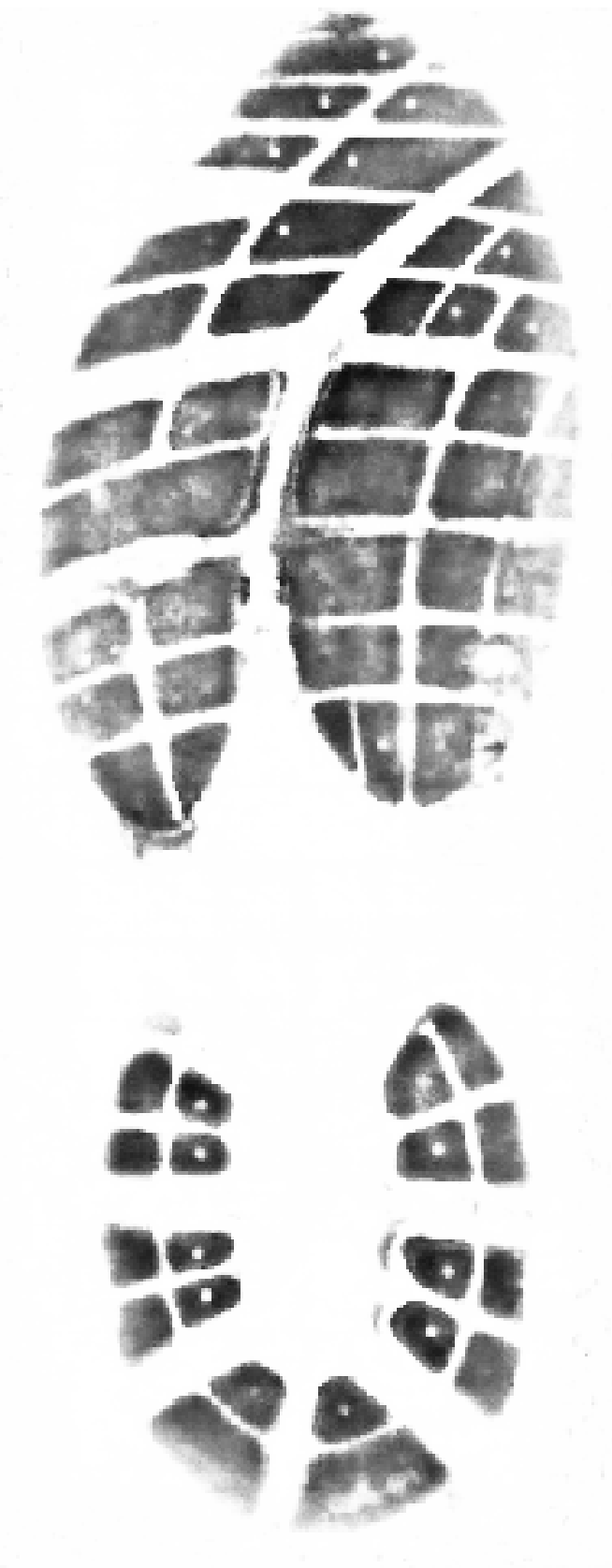}
      \caption{$\Delta t = 34$.}
      \label{fig:34-pred}
  \end{subfigure}
  \caption{Predictions of the model described in \ref{ssec:forward}, given week 48 as input and a range of values for $\Delta t$.}\label{fig:predictions}
\end{minipage}
\end{figure*}

\section{Methodology}
For the task of modelling wear patterns, we implement a CNN architecture in the style of an auto-encoder, inspired by the work in Tatarchenko et al. \cite{tatarchenko2016multi} and Vukotić et al. \cite{vukotic2017onestep}. This architecture consists of three branches---an encoder $F_\theta(\cdot)$ that takes as input a shoeprint image $X$, a delta branch $G_\omega(\cdot)$ that encodes a representation of time from a parameter $\Delta_t$, and a decoder $H_\beta(\cdot)$ that learns an upsampling function to predict the wear pattern. These branches take the form shown in \eqref{eq:architecture}. 

\begin{multline}\label{eq:architecture}
  F_\theta = \sigma(X * \theta) \equiv f,\\
  G_\omega = \sigma(\Delta_t \cdot \omega) \equiv g,\\
  H_\beta = \sigma(f \concat g *' \beta)
\end{multline}

where $\sigma$ represents an activation function---ReLU or sigmoid, $*$ represents the convolution operation, $\concat$ the concatenation of two tensors, and $*'$ the transpose convolution. Bias terms are omitted for notational convenience. 

The encoder is made up of 5 convolutional layers that act as feature extractors by performing discrete convolutions over the input image, with an increasing depth of feature maps. We double feature maps with each layer, going from 32 in the first layer, to 512 in the last convolutional layer. The delta branch consists of 2 fully connected layers; the output of this branch is reshaped and concatenated with the output of the last convolutional layer. This tensor is then fed into the 5 transpose convolutional layers of the decoder, which successively upsample the extracted feature maps and the output of the delta branch to produce an output of the same dimensions as the input. 

Transpose convolutional layers are used here as a learnable upsampling function, as opposed to a fixed upsampling function (such as bilinear) in combination with 2D convolutions, as frequently seen in the literature. We discard pooling layers as traditionally seen in convolutional architectures since our denoising method removes redundant information in the image in the pre-processing stage. 

The parameters of the network are updated by minimising the squared error loss:

\begin{equation}\label{eq:mse}
  L = \frac{1}{n}\sum_{i=1}^{n} \left || H_\beta( F_\theta(X_i) \concat G_\omega(\Delta_t)) - Y_i \right || ^2 _2
\end{equation}

where $n$ is the number of training images presented to the network in one epoch, $X_i \in \mathbb{R} ^2$ is the $i$th image, and $Y_i$ the ground-truth image that corresponds to the input.

A visualisation of this architecture is shown in Figure \ref{fig:arch}.

\begin{figure*}[ht]
  \centering
  \begin{minipage}[t]{\textwidth}
  \begin{subfigure}[t]{0.3\textwidth}
      \includegraphics[width=\textwidth]{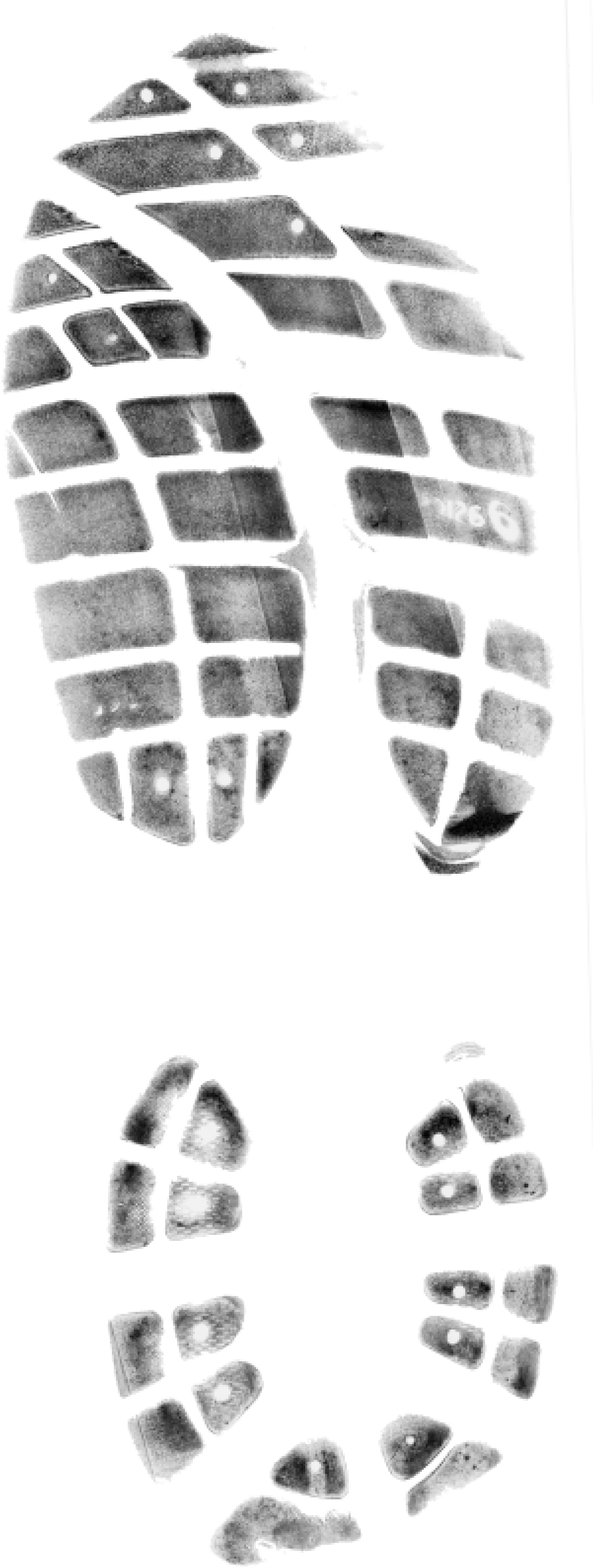}
      \caption{Input image of right outsole on week 42.}
      \label{fig:42R-input}
  \end{subfigure}
  \hfill
  \begin{subfigure}[t]{0.3\textwidth}
      \includegraphics[width=\textwidth]{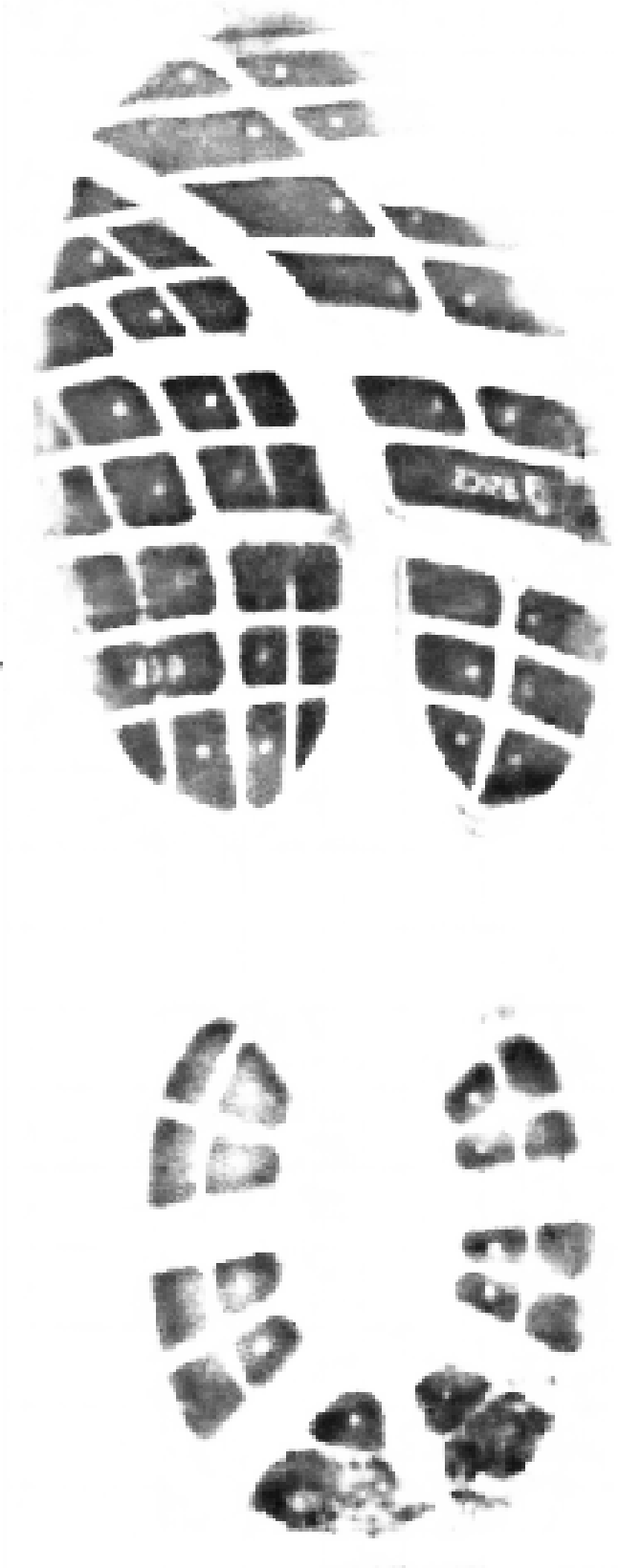}
      \caption{Model prediction given $\Delta t = 20$.}
      \label{fig:20-recon}
  \end{subfigure}
  \hfill
  \begin{subfigure}[t]{0.3\textwidth}
    \includegraphics[width=\textwidth]{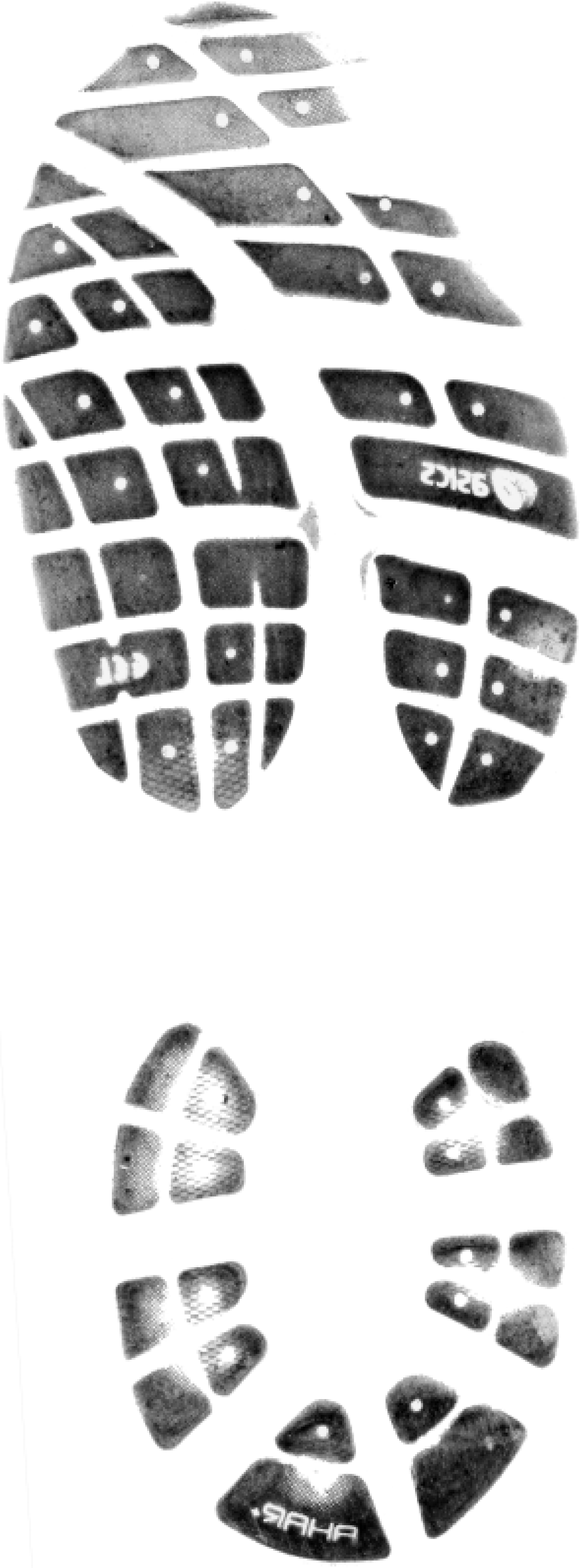}
    \caption{Ground truth image of right outsole on week 20.}
    \label{fig:20R-gt}
  \end{subfigure}
  \caption{Outsole reconstruction predicted by the model detailed in \ref{ssec:backward}.}\label{fig:reconstructions}
\end{minipage}
\end{figure*}

\subsection{Moving Forward: Outsole Wear Prediction}
\label{ssec:forward}
Our first model is designed to extrapolate wear patterns present in the shoeprint and form a prediction of what they might look like after a given period of time, denoted by $\Delta t$. The input image is presented at current relative time $t_0 = 0$. We train this model to predict the appearance of the input, after the elapsed time $\Delta t$, where $\Delta t \in [0, 52]$. $\Delta t$ is incremented in steps of 2 to maintain consistency with the timeframe captured in our dataset. The model then predicts the shoeprint at $t_0 + \Delta t$. 

Formally, we train the model by feeding inputs as batches of $\{X, \Delta t, Y\}$ tuples, where $X$ represents the input image centered at a current relative time $t_0 = 0$; $\Delta t$ represents the desired temporal displacement; and $Y$ represents the ground truth shoeprint image after the desired temporal displacement. Figure \ref{fig:predictions} shows a sample of predictions from this model.

\subsection{Moving Backward: Outsole Reconstruction}
\label{ssec:backward}
Our second model is one that reconstructs the input shoeprint back to its state on any given week in a timeframe of one year. For this task, we use the same architecture as in \ref{ssec:forward}. The only difference with this model is in how we design the $\Delta t$ parameter. Here, $\Delta t$ is represented as a logical vector $\in \mathbb{R}^{52\times1}$; wherein each element represents a week of the year, taking a value in $\{0, 1\}$, such that the desired week corresponding to the ground truth $Y$ is represented as $1$, and all other weeks represented as $0$. 

Once again, we train the model by presenting $\{X, \Delta t, Y\}$ tuples, and design the logical $\Delta t$ vector in increments of $2$ to correspond with the fortnightly nature of our captured dataset. Figure \ref{fig:20-recon} shows a reconstruction produced by this model.

\begin{figure*}[ht]
  \centering
  \begin{minipage}[t]{\textwidth}

  \begin{subfigure}[t]{\textwidth}
    \includegraphics[width=0.18\textwidth]{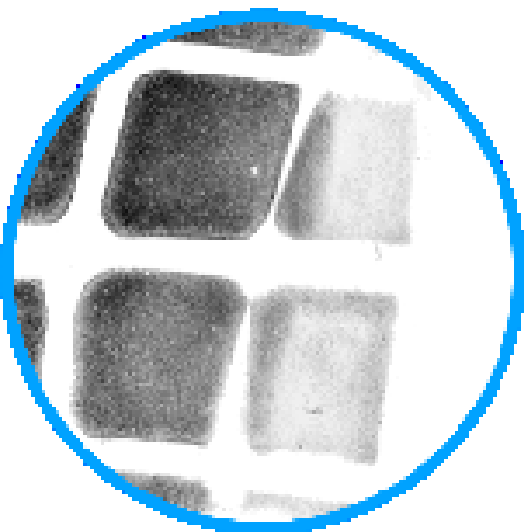}\hfill
    \includegraphics[width=0.18\textwidth]{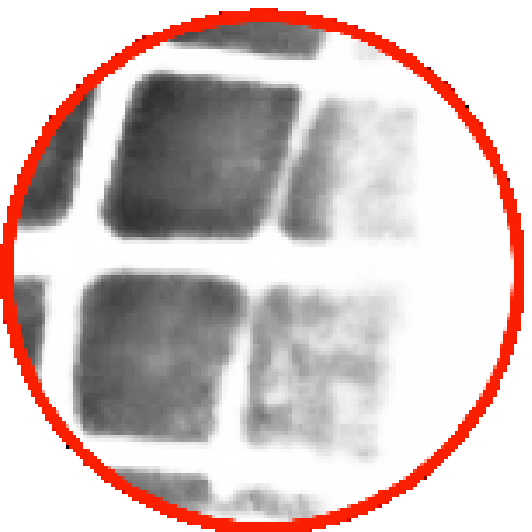}\hfill
    \includegraphics[width=0.18\textwidth]{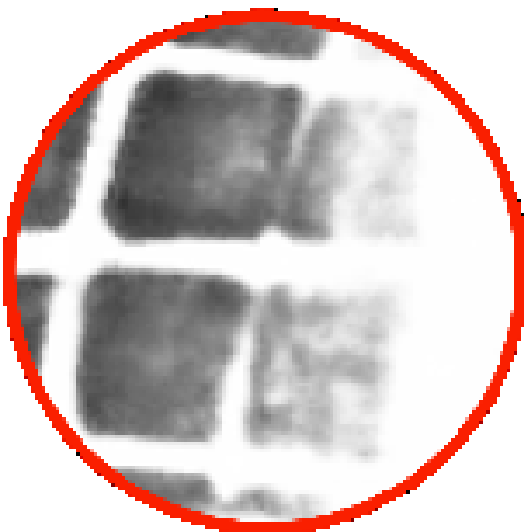}\hfill
    \includegraphics[width=0.18\textwidth]{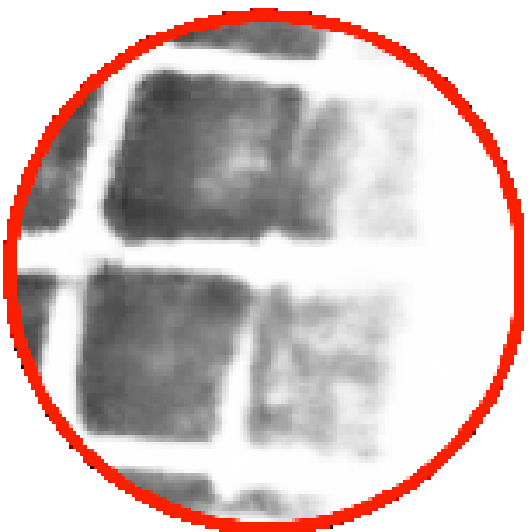}\hfill
    \caption{Cluster of block features and their wear predicted by model.}
    \label{fig:pred-block}
  \end{subfigure}

  \begin{subfigure}[t]{\textwidth}
    \includegraphics[width=0.18\textwidth]{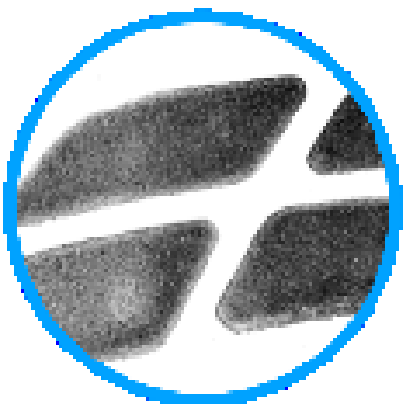}\hfill
    \includegraphics[width=0.18\textwidth]{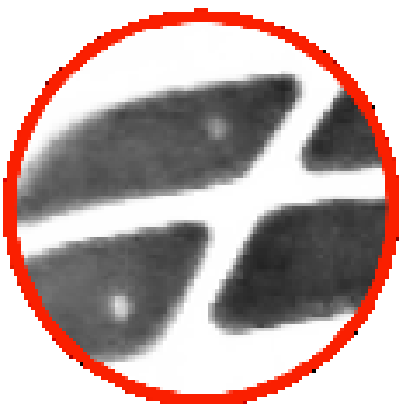}\hfill
    \includegraphics[width=0.18\textwidth]{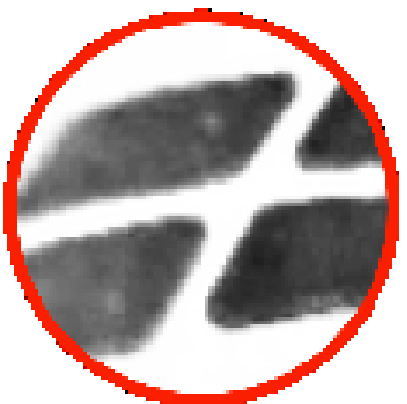}\hfill
    \includegraphics[width=0.18\textwidth]{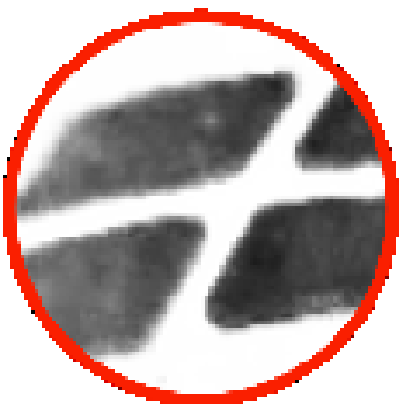}\hfill
    \caption{Dot feature predicted by model.}
    \label{fig:pred-dot}
  \end{subfigure}

  \begin{subfigure}[t]{\textwidth}
    \includegraphics[width=0.18\textwidth]{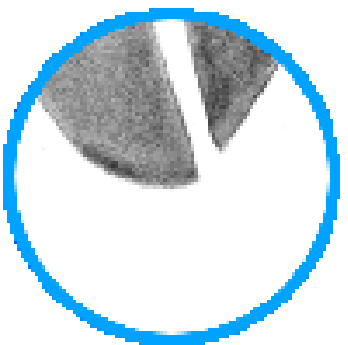}\hfill
    \includegraphics[width=0.18\textwidth]{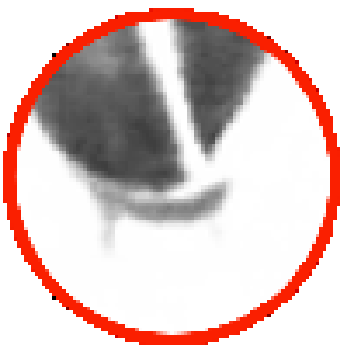}\hfill
    \includegraphics[width=0.18\textwidth]{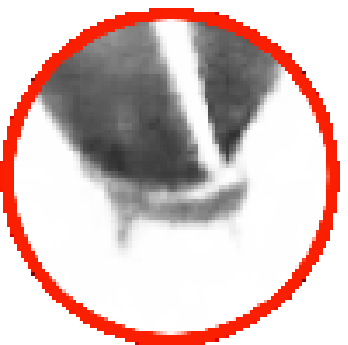}\hfill
    \includegraphics[width=0.18\textwidth]{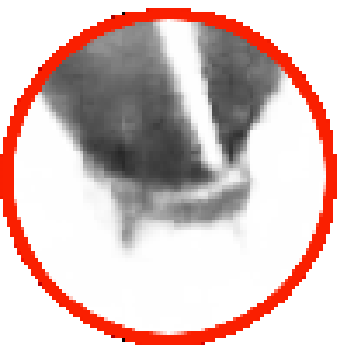}\hfill
    \caption{Outsole feature predicted by model.}
    \label{fig:pred-feature}
  \end{subfigure}

  \caption{Highlights of relevant regions of predictions shown in Figure \ref{fig:predictions}. Encircled in blue is the input to the model of the left outsole on week 48. Circled in red are the predicted wear patterns of the model, given $\Delta_t$ values of 8, 24, and 34, respectively from left to right.}\label{fig:preds-crop}
  \end{minipage}
\end{figure*}

\section{Experiments}
\label{sec:exp}

\subsection{Parameters}
Model training is performed by dividing the dataset of 52 images into an 80/20 training/test split. The first 42 images --- both left and right outsoles --- are used to train the model in conjunction with a MSE loss and the Adam optimisation algorithm \cite{kingma2014adam}. For training model \ref{ssec:backward}, the split is reversed --- i.e. we use the last 42 images for training, and test with the remaining images in our dataset that capture the start of our timeline. We use a learning rate of $1e-5$ and train for 10,000 epochs. Activation functions throughout are the ReLU; except for the last layer which uses a sigmoid function to obtain outputs $\in [0, 1]$. We found the same hyperparameters to be effective for both our models. 

Our dataset is composed of 52 grayscale images with a resolution of 13750$\times$5500. Fitting this dataset into memory during training required downsampling it to 640$\times$256. We train both models end-to-end from random initialisation, to generate the desired output outsole given the image of a shoeprint from our dataset. Alternative learning rates, initialisation schemes, optimisers, and loss functions were evaluated before settling on the above. 

\begin{figure*}
  \centering
  \begin{minipage}[t]{\textwidth}

  \begin{subfigure}[t]{.5\textwidth}
      \includegraphics[width=0.32\textwidth]{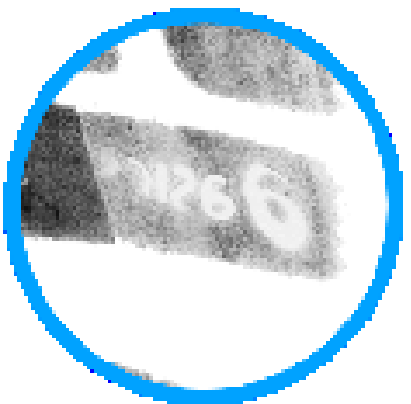}
      \includegraphics[width=0.32\textwidth]{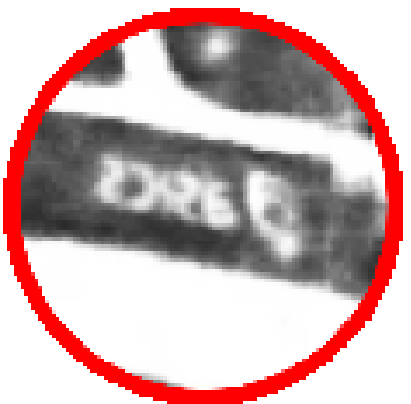}
      \includegraphics[width=0.32\textwidth]{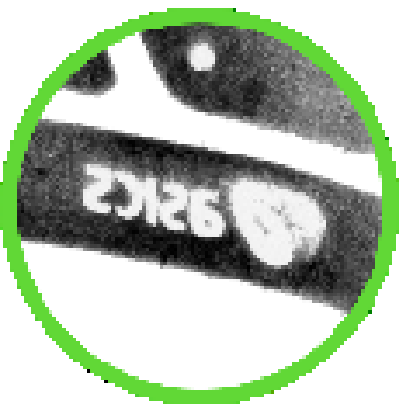}
      \caption{\textit{Asics} brand logo.}
      \label{fig:recon-asics}
  \end{subfigure}
  \hfill
  \begin{subfigure}[t]{.5\textwidth}
      \includegraphics[width=0.32\textwidth]{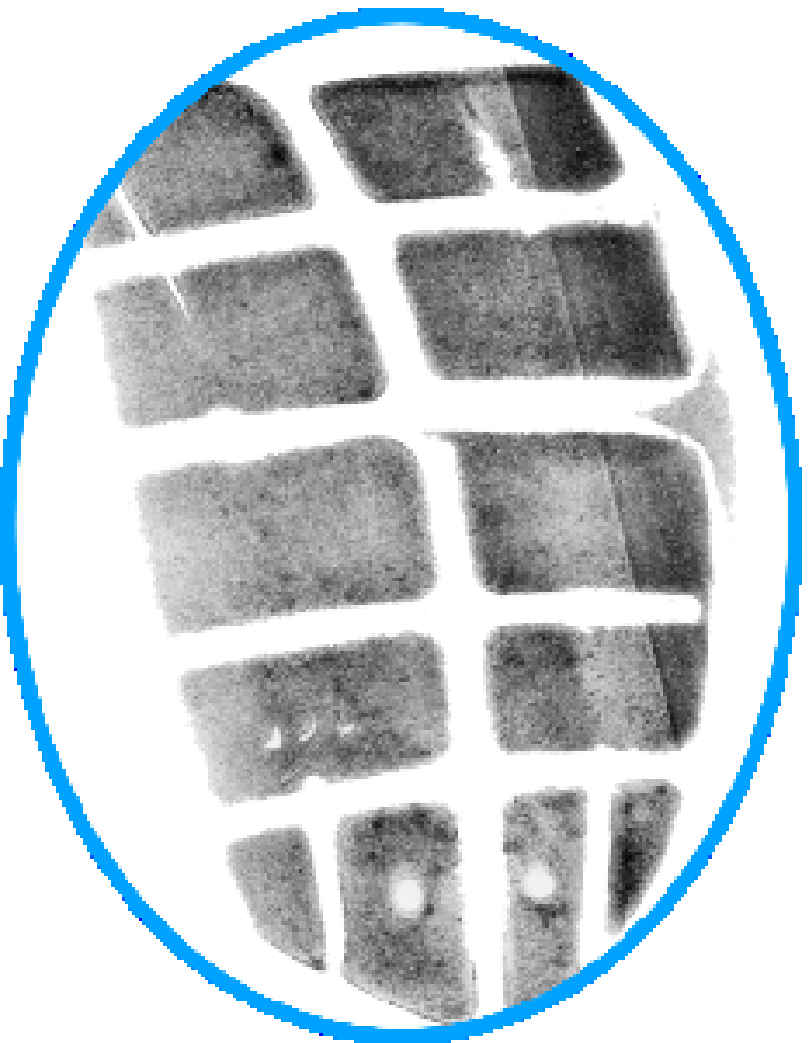}
      \includegraphics[width=0.32\textwidth]{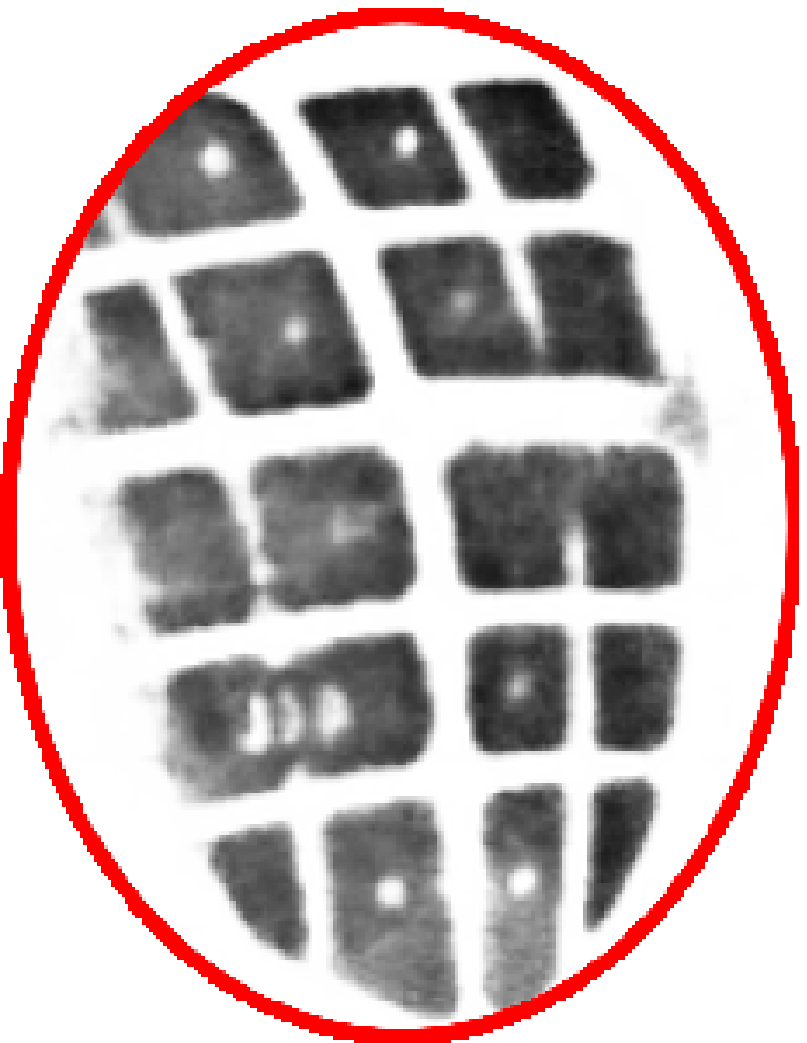}
      \includegraphics[width=0.32\textwidth]{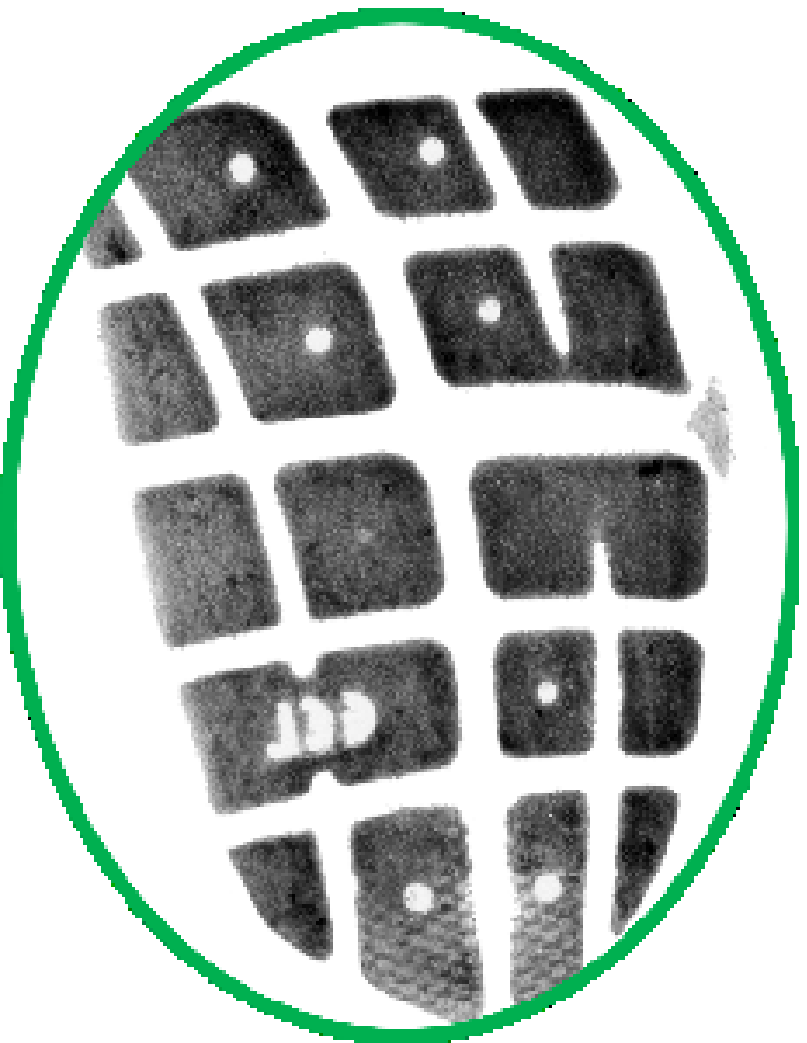}
      \caption{Block and `GEL' features on the outsole.}
      \label{fig:recon-block}
  \end{subfigure}

  \caption{Highlights of relevant regions of outsoles shown in Figure \ref{fig:reconstructions}. Encircled in blue is the input to the model of the right outsole on week 42. Circled in red is the predicted reconstruction of the model for week 20, and circled in green is the ground-truth image of week 20 from the dataset.}\label{fig:recons-crop}
  \end{minipage}
\end{figure*}

\subsection{Results}
Our network successfully learns to model the high-level wear pattern embedded in the shoeprints. From observing the outputs, it is evident that the model has formed an internal representation sufficiently capable of predicting the wear pattern found in the dataset. Relevant regions of Figure \ref{fig:predictions} have been cropped and highlighted in Figure \ref{fig:preds-crop}. Similarly, Figure \ref{fig:recons-crop} consists of crops of Figure \ref{fig:reconstructions}. We compare the predictions from the model against the ground-truth images from our dataset and note the below observations:

\begin{itemize}
  \item In Figure \ref{fig:pred-block} we see a cluster of four block features on the right edge of the outsole. In the model's predictions, we see them degrade and eventually merge in the final prediction, $\Delta_t = 34$. From the ground-truth image of week 52, we confirm that this change has indeed occurred; although clearly the model's estimation of 20 weeks is far off from the reality of this eventual merger materialising in 4 weeks.
  
  \item Figure \ref{fig:pred-dot} shows two `dot features' visible in the first prediction of $\Delta_t = 8$; note that these two features are not present in the input image of week 46. This feature is present throughout the outsole on many of the block features but have disappeared through wear-and-tear. It also happens to be visible in this exact region in all of the training images---weeks 0 through 42---but had eroded from the outsole by the time the input shoeprint was captured. 
  
  Interestingly, in the model's latter two predictions---$\Delta_t = 24, 34$---we see the feature degrade and eventually disappear, in line with the ground-truth; showing that the model has learned the wear development on this and similar features, despite consistently observing the dot features in this region throughout the training images.

  \item Figure \ref{fig:pred-feature} highlights a `ridge' feature seen in all the predictions, but not in the input. We verify through our dataset that this is in fact a feature of the outsole, seen in roughly half of the images in the training set, but is missing in the input image. Note how the model's predictions show this ridge growing progressively larger in size, as the outsole erodes. 
  
  \item In the outsole reconstructions of model \ref{ssec:backward}, we see the successful reproduction of the \textit{Asics} brand logo (Figure \ref{fig:recon-asics}), and the separation of block features that had merged through wear (Figure \ref{fig:recon-block}). Also note the reconstruction of the feature that spells the word `GEL', imprinted by the manufacturer.
  
  \item The bottom region of the heel in the prediction seen in Figure \ref{fig:20-recon} is blurry and poorly defined. This is due to the inconsistency of the appearance of this region in the dataset. During the data collection phase this region was either frequently occluded by fingerprints and debris, or ill-formed due to a lack of pressure between the outsole and the gel while collecting the impression. We deduce that this inconsistency in appearance is what has led the model to develop a fuzzy representation of this region. 
\end{itemize}

From our evaluation of the results, we ascertain that our methodology is sufficient to capture the wear pattern from our dataset, and to perform both outsole prediction and reconstruction with reasonable accuracy. 

We also note that our network is handicapped by a lack of training data. In the era of deep learning, where models are routinely trained with millions of datapoints, we have sufficed with a meagre 52 images. Despite the size of the dataset, each pixel in the input image is a feature the model can learn from, and the 640 $\times$ 256 resolution of our training data is purely limited by processing power; allowing for a more robust model to be trained using higher resolution images. The generalisation ability of deep learning models can also benefit from an adequately sized dataset that fully captures the diversity of the problem domain. 

Empirical evaluations are given in the next subsection. 

\begin{table}[h]
  \centering
  \caption{Mean and standard deviation of SSIM scores.}
  \resizebox{!}{!}{%
  \begin{tabular}{|c|c|c|}
      \hline
      \textbf{} & \textbf{Model \ref{ssec:forward}} & \textbf{Model \ref{ssec:backward}} \\ \hline
      \textbf{Mean} & 0.8645 & 0.8596 \\ \hline
      \textbf{STD} & 0.0381 & 0.0345 \\ \hline
  \end{tabular}%
  }
\label{table:evalssim}
\end{table}

\subsection{Evaluation}
For an objective evaluation of the performance of our models, we use the standard metric of \textit{Structural Similarity Index} (SSIM) \cite{WangSSIM2004}, by comparing the predictions of the models against the ground-truth images from the validation dataset. SSIM is defined in \eqref{eq:ssim}.

\begin{equation}\label{eq:ssim}
    SSIM(f,g) = l(f,g)c(f,g)s(f,g)
\end{equation}

where

\begin{displaymath}
    \begin{cases}˘
        l(f,g) = \frac{2\mu_f\mu_g+C_1}{\mu_f^2+\mu_g^2+C_1},\\
        c(f,g) = \frac{2\sigma_f\sigma_g+C_2}{\sigma_f^2+\sigma_g^2+C_2},\\
        s(f,g) = \frac{\sigma_{fg}+C_3}{\sigma_f\sigma_g+C_3}
    \end{cases}
\end{displaymath}

where $l$, $c$, and $s$ denote the luminance, contrast, and structure comparison functions respectively. The term $f$ denotes the ground-truth image and $g$ the predicted image. $\mu$ and $\sigma$ denote mean and standard deviation of image luminance and contrast, respectively. $\sigma_{fg}$ is the covariance between $f$ and $g$. $C_1$, $C_2$, and $C_3$ are positive constants employed to avoid a null denominator. The SSIM index is a postive value in [0, 1], where 0 denotes no correlation and 1 denotes $f = g$. 

The results are given in Table \ref{table:evalssim}. As evident, the models have an average accuracy of 86\%. 

Additionally, we compared \textit{Peak Signal-to-Noise Ratio} (PSNR) \eqref{eq:psnr} scores for our models.

\begin{equation}\label{eq:psnr}
    PSNR(f,g) = 10\log_{10}(255^{2}/MSE(f,g))
\end{equation}

Once again, in \eqref{eq:psnr} as in \eqref{eq:ssim}, $f$ and $g$ denote the ground-truth and predicted images, respectively; and $MSE$ is the mean squared error --- i.e. $MSE$ between $f$ and $g$. The PSNR score for model \ref{ssec:forward} has an average of 22dB, while model \ref{ssec:backward} scores 21.3dB. Results are given in Table \ref{table:evalpsnr}. 

\begin{table}[h]
  \centering
  \caption{Mean and standard deviation of PSNR scores.}
  \resizebox{!}{!}{%
  \begin{tabular}{|c|c|c|}
      \hline
      \textbf{} & \textbf{Model \ref{ssec:forward}} & \textbf{Model \ref{ssec:backward}} \\ \hline
      \textbf{Mean} & 22.0065 & 21.3681 \\ \hline
      \textbf{STD} & 1.8930 & 1.8820 \\ \hline
  \end{tabular}%
  }
\label{table:evalpsnr}
\end{table}

\section{Conclusion}
\label{sec:conclusion}
We present a convolutional neural network architecture in the style of an auto-encoder that, for the first time, can model the wear pattern collected in a unique dataset of shoeprints. We show that the model can learn an accurate representation of the pattern of wear-and-tear found in the shoeprints by applying it to predict the wear pattern on the outsole after a given temporal displacement; and by having it reconstruct the outsole back to its original state at a previous point in time. We address the drawbacks of the model and present objective evaluations of its performance, which show the predictions of both models to be 86\% accurate.

This work adds to the scant literature on shoeprint wear patterns by presenting a computational model of outsole wear. The model presented within can be applied to supplement the skills and expertise of the forensic examiner in their analysis of crime scene shoeprints; and additionally to train the novice forensic scientist to hone their skills. 

\section*{Acknowledgements}
The authors express their gratitude to the \textit{High Technology Transdisciplinary Research Network} at Unitec, and the \textit{Institute of Environmental Science and Research} (ESR), New Zealand for jointly funding this research.

\bibliographystyle{IEEEbib}
\bibliography{refs}

\end{document}